\documentclass[10pt,twocolumn,letterpaper]{article}
\pdfoutput=1
\usepackage{iccv}
\usepackage{times}
\usepackage{epsfig}
\usepackage{graphicx}
\usepackage{amsmath}
\usepackage{amssymb}

\usepackage{bbm}
\usepackage{multirow} 
\usepackage{tabularx}
\usepackage{stfloats}
\usepackage{xcolor}
\usepackage{diagbox}
 \usepackage{algorithm}
\usepackage{algorithmic}

\usepackage{dsfont}
 
\usepackage[pagebackref=true,breaklinks=true,letterpaper=true,colorlinks,bookmarks=false]{hyperref}

\iccvfinalcopy 

\def\iccvPaperID{2136} 

\ificcvfinal\fi

\begin{document}

\title{Semi-supervised Semantic Segmentation Meets Masked Modeling:\\Fine-grained Locality Learning Matters in Consistency Regularization}

\author{
Wentao Pan$^{1,2}$\footnotemark[1], 
Zhe Xu$^{3}$, 
Jiangpeng Yan$^{1}$, 
Zihan Wu$^{2}$, 
Raymond Kai-yu Tong$^{3}$, 
Xiu Li$^{1}$\footnotemark[2] \ , 
Jianhua Yao$^{2}$\footnotemark[2]
\\ \\
$^{1}$Tsinghua Shenzhen International Graduate School, Tsinghua University \\ $^{2}$Tencent AI Lab  \\  $^{3}$Department of Biomedical Engineering, The Chinese University of Hong Kong\\ \\
}
\maketitle
\renewcommand{\thefootnote}{\fnsymbol{footnote}}
\footnotetext[1]{work done during an intern at Tencent AI Lab. }
\footnotetext[2]{corresponding authors}

\ificcvfinal\thispagestyle{empty}\fi

\begin{abstract}
Semi-supervised semantic segmentation aims to utilize limited labeled images and abundant unlabeled images to achieve label-efficient learning, wherein the weak-to-strong consistency regularization framework, popularized by FixMatch, is widely used as a benchmark scheme. 
Despite its effectiveness, we observe that such scheme struggles with satisfactory segmentation for the local regions. 
This can be because it originally stems from the image classification task and lacks specialized mechanisms to capture fine-grained local semantics that prioritizes in dense prediction. 
To address this issue, we propose a novel framework called \texttt{MaskMatch}, which enables fine-grained locality learning to achieve better dense segmentation. 
On top of the original teacher-student framework, we design a masked modeling proxy task that encourages the student model to predict the segmentation given the unmasked image patches (even with 30\% only) and enforces the predictions to be consistent with pseudo-labels generated by the teacher model using the complete image. 
Such design is motivated by the intuition that if the predictions are more consistent given insufficient neighboring information, stronger fine-grained locality perception is achieved. 
Besides, recognizing the importance of reliable pseudo-labels in the above locality learning and the original consistency learning scheme, we design a multi-scale ensembling strategy that considers context at different levels of abstraction for pseudo-label generation. 
Extensive experiments on benchmark datasets demonstrate the superiority of our method against previous approaches and its plug-and-play flexibility. 
\end{abstract}

\section{Introduction}

Semantic segmentation is a fundamental task in computer vision and is widely used in defect detection, autopilot and computer-aided diagnosis \cite{hao2020brief}.
Nevertheless, existing top-performing methods usually rely on a large scale of pixel-wise labeled data, which is a labor-intensive and time-consuming. 
In this regard, semi-supervised learning (SSL) paradigm has emerged as a solution to this challenge by making use of both limited labeled data and vast amounts of unlabeled data simultaneously.

\begin{figure}[t]
\begin{center}
\includegraphics[width=\linewidth]{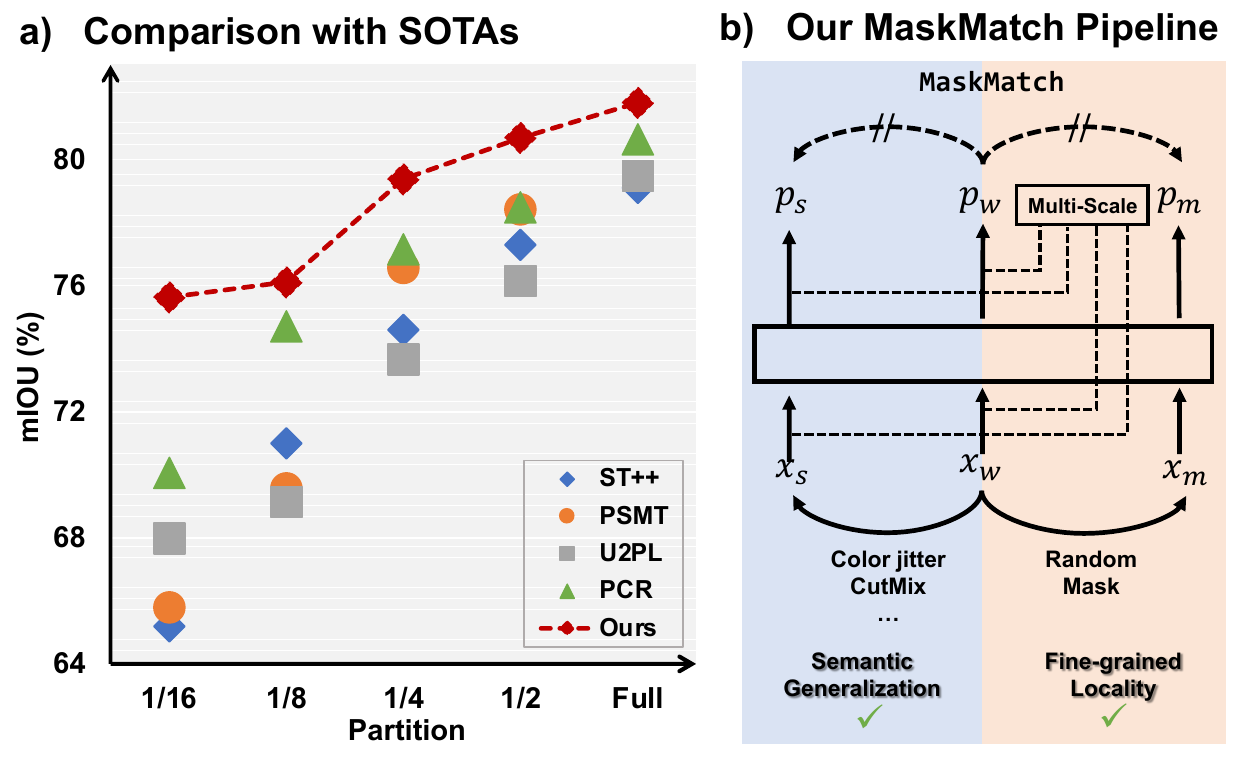}
\end{center}
\vspace{-1.0em}
   \caption{
   a) we compare our method with state-of-the-art methods on \textit{Pascal VOC 2012} under diverse classic partition protocols. 
   Clear performance margins can be observed. b) our MaskMatch consists of two branches. 
   Especially, the right branch is the proposed local consistency regularization powered by the masked modeling scheme.
   A multi-scale ensembling strategy is also incorporated to enhance pseudo-labeling. 
   Both strategies (in orange) help improve the fine-grained locality learning ability upon the original weak-to-strong consistency regularization framework (in blue).
   }
\label{fig:fig1}
\vspace{-1.0em}
\end{figure}

One widely adopted SSL paradigm is the smoothness assumption and low-density assumption \cite{van2020survey} based weak-to-strong consistency regularization framework, which is popularized by FixMatch \cite{sohn2020fixmatch} for image classification. 
This scheme has also shined on semantic segmentation \cite{french2019semi,ouali2020semi,fan2022ucc,liu2022perturbed}, and become a widely used baseline. Generally, it incorporates consistency regularization with self-training, where the one-hot pseudo labels from the prediction of weakly perturbed unlabeled images $x_w$ are utilized to supervise the prediction of strongly perturbed unlabeled images $x_s$, as depicted in Figure \ref{fig:fig1} (b). 
The success of this approach can be attributed to that the model is more inclined to generate accurate predictions on $x_w$. 
Typically, strong perturbations help mitigate confirmation bias and can be performed by image perturbations (e.g., color jitter, CutMix \cite{yun2019cutmix}, context augmentation \cite{lai2021semi}) or feature perturbations (e.g., VAT \cite{liu2022perturbed}). 
Despite effectiveness in leveraging unlabeled images to boost segmentation performance \cite{wang2022semi,fan2022ucc,liu2022perturbed,unimatch}, we observe that such scheme often struggles to achieve satisfactory segmentation results for local regions. 
This may be because this scheme was originally proposed for image-level classification and lacks specialized mechanisms to capture fine-grained local semantics that are necessary for dense pixel-level prediction. 
Therefore, it is natural to ask whether there is an auxiliary proxy task that can enhance the model's perception of local semantics without significantly increasing the number of parameters.

\begin{figure*}[t]
\vspace{-1.0em}
\centering
\includegraphics[width=0.8\linewidth]{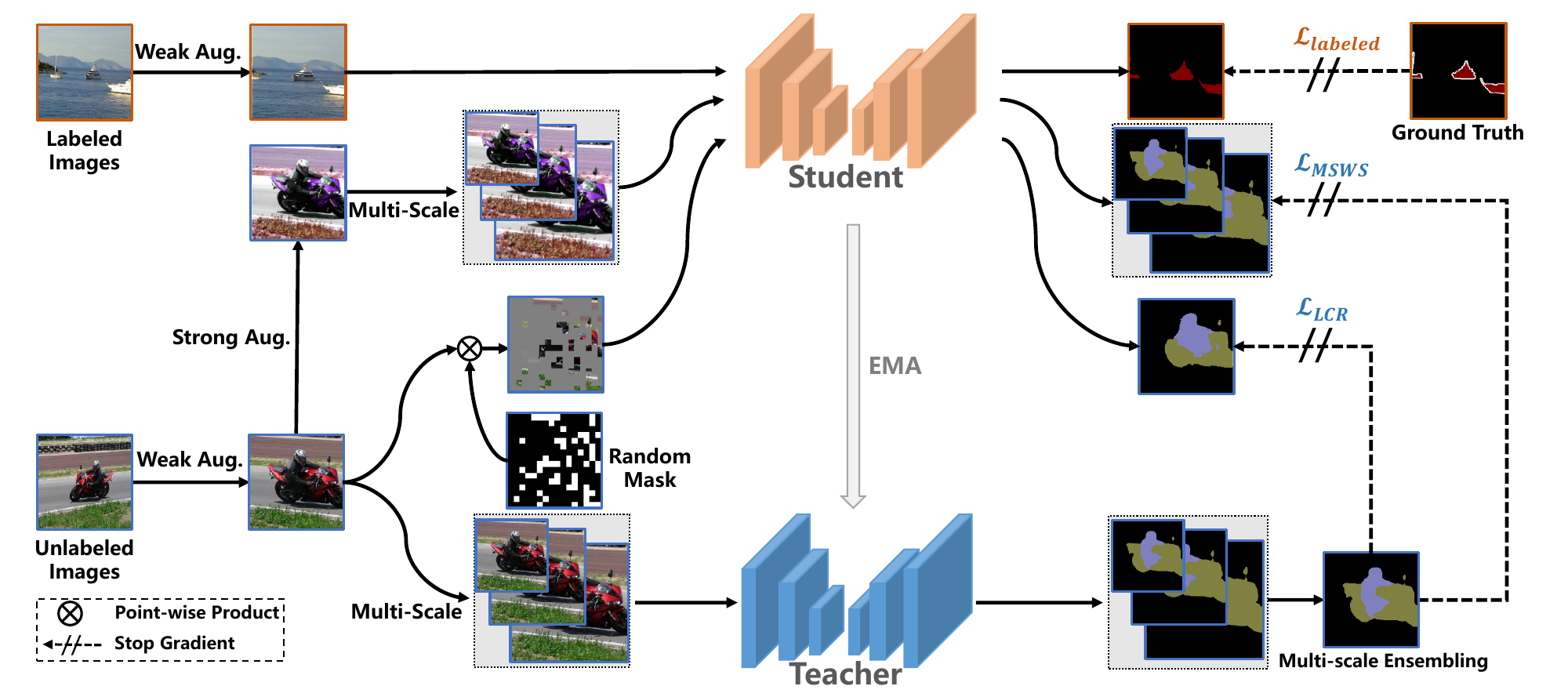}
\caption{
Overview of \texttt{MaskMatch}. The method is built on the mean-teacher paradigm, where the weights of the teacher model are updated by the exponential moving average (EMA) weights of the in-training student model. 
For \textbf{local consistency regularization (LCR)}, randomly masked weakly-augmented unlabeled images are fed to the student model to obtain the entire segmentation prediction.
For \textbf{multi-scale enhanced consistency (MS)}, multi-scale transformations are applied on both strongly and weakly augmented unlabeled images, and fed to the student and teacher models separately.
The pseudo labels for the above two consistency terms are the average of teacher's predictions of multi-scale weakly-augmented unlabeled images.
}
\label{fig:fig2}
\end{figure*}

Towards the above question, in this paper, we propose a novel semi-supervised segmentation framework called \texttt{MaskMatch} based on masked modeling and multi-scale pseudo-labeling, stepping into fine-grained locality learning to achieve better dense prediction. 
Figure \ref{fig:fig1} (b) compares our method with the previous approach. On top of the original teacher-student framework, we design an auxiliary masked modeling proxy task towards learning locality-enhanced semantic representation, as shown in Figure \ref{fig:fig2}. 
Such design shares similar spirit with masked image modeling (MIM) pre-training \cite{he2022masked, xie2022simmim}, which randomly masks several patches in images (even with 75\% masking ratio \cite{he2022masked}) and reconstructs the entire image using the rest unmasked patches. 
Yet, unlike MIM, which is powered by the self-supervised low-level image reconstruction task, one interesting and effective design here is that our proxy is forced to directly perform the challenging high-level segmentation task to provide more task-specific guidance (as demonstrated in Table \ref{tab:dissusion_task}). 
Specifically, the tailored proxy encourages the student model to make associations and perform complete segmentation given the unmasked image patches only (as visualized in Figure \ref{fig:mask_sampling_strategies}), and then enforces the predictions to be consistent with pseudo-labels generated by the teacher model using the complete image. 
Intuitively, given only a few neighboring patch clues, more consistent predictions indicate stronger capability of fine-grained semantics perception since the model is required to better perform implicit low-level reconstruction of the masked local regions and subsequent high-level object segmentation. 
Furthermore, recognizing the importance of reliable pseudo-labels in our locality learning and the original consistency learning, besides the typical confidence thresholding, we introduce a multi-scale ensembling strategy that considers context at different levels of abstraction to further improve the quality of pseudo labels. 
In summary, our contributions are mainly three-fold:

\begin{itemize}
    \item
    We propose a novel semi-supervised semantic segmentation framework which incorporates masked modeling for fine-grained locality learning. 
    To our best knowledge, we are the first to tailor the spirit of masked modeling into semi-supervised segmentation with interesting findings that task-specific modeling is more beneficial (Table \ref{tab:dissusion_task}).
  
    \item 
    We propose a multi-scale ensembling strategy for more comprehensive pseudo-label generation, which further enhances the locality learning ability of the original weak-to-strong consistency learning.
    
    \item 
    Extensive experiments on two public benchmark datasets show the superior performance of our method against previous approaches with a large margin (as depicted in Figure \ref{fig:fig1} (a)). 
    Noteworthily, the proposed strategies do not introduce extra parameters or modify the backbone, ensuring plug-and-play flexibility. 
\end{itemize}

\section{Related Work}

\noindent\textbf{Semi-Supervised Learning.} 
Semi-supervised learning is widely studied in image classification.
Mainstream methods fall into two categories: self-training and consistency regularization. The self-training (also known as Pseudo-labeling) approaches \cite{lee2013pseudo, arazo2020pseudo} utilize unlabeled images in a supervised-like manner with one-hot pseudo labels generated by the up-to-date optimized model itself. 
Consistency regularization methods \cite{tarvainen2017mean, miyato2018virtual, xie2020unsupervised} improve the generalizability by enforcing the consistency among the predictions of unlabeled images with perturbations. 
Recent works have shown that self-training and consistency regularization can be used together. 
The popular FixMatch \cite{sohn2020fixmatch}, for instance, employs predictions of weakly perturbed unlabeled images as pseudo labels to constrain the prediction of strongly perturbed unlabeled images.

\noindent\textbf{Semi-Supervised Semantic Segmentation.} 
Borrowing the spirit of semi-supervised image classification, self-training and consistency regularization also shines on the segmentation task \cite{ouali2020semi, french2019semi, ke2020guided, zou2020pseudoseg, chen2021semi, hu2021semi, yang2022st++, liu2022perturbed, xu2023ambiguity, wang2022semi, fan2022ucc, xu2022semi}. 
Likewise, these approaches incorporate both consistency regularization and self-training into the mean-teacher framework \cite{tarvainen2017mean} or the co-teaching paradigm \cite{chen2021semi}. 
One line of recent studies focus on effective perturbations, considering that they are crucial in such learning schemes. 
Specifically, input perturbations such as Gaussian noise, color jitter, CutOut \cite{devries2017improved}, CutMix \cite{yun2019cutmix}, and other image augmentation techniques are widely used \cite{hu2021semi, yang2022st++, liu2022perturbed, wang2022semi, fan2022ucc, xu2022semi, unimatch}. 
\cite{liu2022perturbed} and \cite{unimatch} inject feature-level perturbations by virtual adversarial training \cite{Miyato2019VAT} and channel-wise DropOut. 
Another line is concerned about the errors in the pseudo labels, which can easily lead to confirmation bias. 
For example, pseudo label selection \cite{wang2022semi, yang2022st++, kwon2022semi, unimatch} and loss calibration \cite{hu2021semi, liu2022perturbed}. Apart from them, contrastive learning \cite{liu2022reco, zhou2021c3, wang2022semi} and prototype learning \cite{xu2022semi} are also incorporated in the consistency regularization paradigm to achieve discriminative representation. 
However, these methods overlook that consistency regularization was originally proposed for image-level classification and do not have specialized mechanisms for capturing fine-grained local semantics that are critical for dense prediction, which is the main focus of this paper.

\noindent\textbf{Masked Image Modeling.} 
Masked language modeling has been a successful technique for language representation learning as seen in the case of BERT \cite{bert}. 
To apply this idea to vision representation learning, masked image modeling (MIM) was proposed, where random patches of an image are masked, and the information from unmasked patches is used to reconstruct the masked ones \cite{xie2022simmim, he2022masked}. 
While the auxiliary masked modeling proxy task in our \texttt{MaskMatch} shares similar technical spirits with MIM, they differ in both purpose and design. 
MIM is designed for self-supervised representation learning, but ours is for fine-grained locality learning in semi-supervised learning. 
Besides, MIM reconstructs masked patches using unmasked patch clues, but \texttt{MaskMatch} encourages the model to make associations and directly predict the complete segmentation given the unmasked patches only. 
We found that this task-specific design is particularly well-suited for semantic segmentation. 
Recently, MIM and its variants have shown promising results in various downstream vision tasks, including image classification \cite{cai2022semi}, domain adaptation \cite{hoyer2022mic}, knowledge distillation\cite{yang2022masked}, and multi-modal learning \cite{bachmann2022multimae, tong2022videomae}. 
Here we further extend this idea to achieve fine-grained locality perception in semi-supervised segmentation.

\section{Method}

\subsection{Preliminaries and Baseline Setting}
\label{sec:baseline}

The semi-supervised semantic segmentation problem includes two datasets, i.e., a labeled set $D_l = \{(x_i^l, y_i^l) | x_i^l \in \mathbb{R}^{(H, W, 3)}, y_i^l \in \{0,1\}^{(H, W, C)}, 1 \leq i \leq M\}$ containing $M$ samples and an unlabeled set $D_u = \{x_i^u | x_i^u \in \mathbb{R}^{(H, W, 3)}, 1 \leq i \leq N\}$ containing $N$ samples, where 
$x_i^l/x_i^u$ denote $i$-th image in labeled/unlabeled set, $y_i^l$ denotes the corresponding pixel-wise ground truth. 
The number of semantic categories, height, and width of the labeled image $x^l$ are denoted as $C$, $H$, and $W$, respectively.
The goal is to improve the segmentation ability of the model by leveraging $D_l$ and $D_u$ together.

We set up our baseline upon the popular mean-teacher paradigm, which consists of a student model $f(x; \theta)$ and a teacher model $f(x; \theta^{'})$.
The student model undergoes end-to-end training, and the teacher model's weights $\theta^{'}$ are the exponential moving average (EMA) of the student model's weights $\theta$, i.e., $\theta^{'}_{t} = \alpha\theta^{'}_{t-1} + (1-\alpha) \theta_t$ at time step $t$.
The smoothing coefficient $\alpha$ is 0.999 \cite{tarvainen2017mean}.
The labeled images are utilized in the typical supervised manner with cross-entropy loss $\ell_{ce}$, formulated as:
\begin{equation}
\mathcal{L}_{labeled} = \frac{1}{M} \sum_{i=1}^{M} \frac{1}{H W} \sum_{j=1}^{H W} \ell_{ce}(f(x_{i,j}^l;\theta), y_{i,j}^l).
\label{eq:loss_l}
\end{equation}
Then, weak-to-strong consistency regularization is adapted to explore unlabeled images, which constrains the consistency between the predictions of weakly perturbed images $x^{u}_w$ and strongly perturbed images $x^{u}_s$.
Specifically, the weak augmentation $\mathcal{A}_w(\cdot)$ and strong augmentation $\mathcal{A}_s(\cdot)$ are applied on $x^u$ to obtain $x^{u}_w$ and $x^{u}_s$ by:
\begin{equation}
x^{u}_w=\mathcal{A}_w(x^u), \ \ \ \ x^{u}_s=\mathcal{A}_s(\mathcal{A}_w(x^u)).
\label{eq:perturbation}
\end{equation}
The teacher model $f(x; \theta^{'})$ takes $x^{u}_w$ as the input and generates one-hot pseudo-label $\hat{y}^u \in \{0, 1\}^{(H,W,C)}$ based on its prediction. $\hat{y}^u$ is used to supervise the student's prediction of $x^{u}_s$.
As the pseudo labels are inevitably inaccurate, directly utilizing them for supervision will lead the model to learn incorrect information.
Therefore, we discard pseudo-label pixel-by-pixel with a class-wise maximum predicted probability $p^u = \underset{c}{max}(f(\mathcal{A}_w(x_{i}^u); \theta^{'})) \in [0, 1]^{(H, W)}$ higher than a pre-defined threshold $\tau$ during training.
The loss for unlabeled images is defined as: 
\begin{equation}
\begin{aligned}
\mathcal{L}_{unlabeled} & =  \frac{1}{N} \sum_{i=1}^{N} \frac{1}{\sum_{j=1}^{H W} \mathds{1}(p_{i,j}^u > \tau)} \sum_{j=1}^{H W} \\ 
& \mathds{1}(p_{i,j}^u > \tau) \ell_{ce}(f(\mathcal{A}_s(\mathcal{A}_w(x_{i,j}^u));\theta), \hat{y}_{i,j}^u),
\end{aligned}
\label{eq:loss_u_conf}
\end{equation}
where $\mathds{1}(\cdot)$ is the indicator function.

As such, the overall objective $\mathcal{L}$ consists of labeled term $\mathcal{L}_{labeled}$ and unlabeled term $\mathcal{L}_{unlabeled}$, formulated as:
\begin{equation}
\mathcal{L} = \mathcal{L}_{labeled} + \lambda\mathcal{L}_{unlabeled},
\label{eq:loss}
\end{equation}
where $\lambda$ is a trade-off weight.

\subsection{MaskMatch} \label{sec:maskmatch}
The weak-to-strong consistency regularization improves generalization by alleviating confirmation bias with various strong augmentations and constraining the consistency.
However, it often struggles to achieve satisfactory segmentation results for local regions.
This could be attributed to its originally proposed for image-level classification, which lacks specialized mechanisms for capturing fine-grained local semantics that prioritizes in dense pixel-level prediction.
Recently, masked image modeling has shown promising progresses in self-supervised vision representation learning \cite{he2022masked, xie2022simmim}, where the trained model can successfully restore 75\% masked patches given the rest 25\% unmasked patches' information only \cite{he2022masked}, indicating that this strategy forces the models to capture local patterns.  
Partly building on this insight, we tailor a local consistency regularization (LCR), powered by an auxiliary masked modeling proxy task to promote fine-grained locality learning. 
Additionally, we introduce a multi-scale ensembling strategy to generate more reliable pseudo-labels for our LCR and the original consistency regularization.

\noindent\textbf{Local Consistency Regularization (LCR).}
To generate a mask $z^u \in \{0, 1\}^{(H, W)}$ (0/1 indicate masked/unmasked pixels) for the masked modeling proxy task, we start by initializing $z^u$ to be $\{1\}^{(H, W)}$. 
Then, we split image $z^u$ into several $p \times p$ patches without overlap. Each patch is randomly masked with a probability of $\gamma$, resulting in a binary mask indicating which pixels are masked and which ones are unmasked.  The masking process can be formulated as:
\begin{equation}
z^u = \mathcal{M}(x^u; p, \gamma).
\label{eq:mask}
\end{equation}
The generated mask $z^u$ is used to create the masked image $x_m^u = \mathcal{A}_w(x^u) \otimes z^u$, and we employ the pseudo labels $\hat{y}^u$ to supervise the student model's predictions for $x_m^u$. 
As incorrect pseudo labels may harm fine-grained locality learning, we apply pseudo label selection (same as Eq. \ref{eq:loss_u_conf}) to exclude low-confidence pseudo labels during training. 
The loss for local consistency regularization is formulated as:
\begin{equation}
\begin{aligned}
\mathcal{L}_{LCR} & =  \frac{1}{N} \sum_{i=1}^{N} \frac{1}{\sum_{j=1}^{H W} \mathds{1}(p_{i,j}^u > \tau)} \sum_{j=1}^{H W} \\ 
& \mathds{1}(p_{i,j}^u > \tau ) \ell_{ce}(f(\mathcal{A}_w(x_{i,j}^u) \otimes z_{i,j}^u;\theta), \hat{y}_{i,j}^u).
\end{aligned}
\label{eq:loss_lcr}
\end{equation}

Distinct from MIM that performs the low-level mask-then-reconstruction task, our LCR facilitates fine-grained locality learning through a task-specific mask-then-predict approach. 
The success of LCR can be attributed to that it encourages the network to make associations and perform complete segmentation based on only a few patch clues. In other words, if the predictions on masked images become more consistent with the pseudo-labels obtained from complete images, the model possesses stronger object perception abilities since it requires the model to better perform implicit low-level reconstruction of the masked local regions and subsequent high-level object segmentation. 
Note that other mask-then-predict strategies are extensively discussed as shown in Table \ref{tab:dissusion_task}, where we found that the current task-specific modeling is more beneficial.

\noindent\textbf{Multi-scale Ensembling Strategy (MS).}
The quality of pseudo labels is crucial in both our local consistency regularization and the original weak-to-strong consistency regularization.
The scale mismatch of objects in labeled images and unlabeled images is part of the reason for erroneous pseudo labels.
To alleviate this, we further employ a multi-scale ensembling strategy to improve the quality of pseudo labels. 
By generating pseudo labels through a consensus of predictions from models with different resolutions, such strategy helps perceive the same objects at different scales, thus alleviating incorrect pseudo labels caused by scale mismatch.
Specifically, two additional inputs $x^u_1$ and $x^u_2$ are obtained via the interpolation operation $\mathcal{S}(x; \sigma)$ with scale factors $\sigma_1 \in (0, 1)$ and $\sigma_2 \in (1, \infty)$. 
Then, the final ensemble pseudo label, further denoted as $\widetilde{y}^{u}$, is obtained by averaging multi-scale predictions as: 
\begin{equation}
\widetilde{y}^{u} = \underset{c}{argmax}(\frac{\mathcal{S}(\sum_{k=0}^{2}f(\mathcal{S}(\mathcal{A}_w(x^u) ; \sigma_k); \theta^{'}) ; \frac{1}{\sigma_k})}{3}).
\label{eq:multi_scale_pseudo_labels}
\end{equation}
As such, the pseudo label $\widetilde{y}^{u}$ is employed to supervise the prediction for the image of each scale in both the original weak-to-strong consistency regularization and our local consistency regularization. The predictions of multiple scales are resized to the original size of the image for computational efficiency. Thus, the weak-to-strong consistency loss can be further modified as:
\begin{equation}
\begin{aligned}
\mathcal{L}_{MSWS} & =  \frac{1}{N} \sum_{i=1}^{N} \frac{1}{3\sum_{j=1}^{H W} \mathds{1}(\widetilde{p}_{i,j}^u > \tau)} \sum_{j=1}^{H W} \sum_{k=0}^{2}\\ 
& \mathds{1}(\widetilde{p}_{i,j}^u > \tau) \ell_{ce}(\mathcal{S}(f(\mathcal{S}(\mathcal{A}_s(x_{i,j}^u) ; \sigma_k) ; \frac{1}{\sigma_k});\theta), \widetilde{y}_{i,j}^u).
\end{aligned}
\label{eq:loss_msws}
\end{equation}
Finally, the overall objective function of our \texttt{MaskMatch} is a combination of three loss terms, i.e., $\mathcal{L}_{labeled}$ (Eq. \ref{eq:loss_l}), $\mathcal{L}_{MSWS}$ (Eq. \ref{eq:loss_msws}) and $\mathcal{L}_{LCR}$ (Eq. \ref{eq:loss_lcr}):
\begin{equation}
\mathcal{L} = \mathcal{L}_{labeled} + \lambda_1\mathcal{L}_{MSWS}+ \lambda_2\mathcal{L}_{LCR},
\label{eq:loss_all}
\end{equation}
where $\lambda_1$ and $\lambda_2$ are the trade-off weights.

\section{Experiments}
\subsection{Setups}

\noindent\textbf{Datasets:} We evaluate our method on two public datasets, \textit{Pascal VOC 2012} \cite{pascal-voc-2012} and \textit{Cityscapes} \cite{hariharan2011semantic}.
\textbf{\textit{Pascal VOC 2012}} is an object-centered dataset for visual object recognition in realistic scenes, which consists of 21 semantic classes (including background) and a total of 1464/1449 finely-annotated images for training/validation. 
Following the common settings \cite{chen2021semi, liu2022perturbed, wang2022semi}, we incorporate SDB \cite{hariharan2011semantic} as our full training set, which contains 10582 training images. 
\noindent\textbf{\textit{Cityscapes}} is an urban street scenes semantic understanding dataset with 30 labeled classes. 
Rare classes are excluded and 19 classes are utilized for evaluation.
The dataset comprises 2975/500 images for training/validation.

\noindent\textbf{Evaluation:} 
We follow the same partitions used in U$^2$PL \cite{wang2022semi} and CPS \cite{chen2021semi} for fair comparison.
During evaluation, the predictions are generated by the teacher model. Sliding window inference is adapted for \textit{Cityscapes}.

\noindent\textbf{Implementation Details:} Following the standard practice, we used DeepLabV3+ \cite{chen2018encoder} with ResNet-50/ResNet-101 \cite{he2016deep} backbone, which is initialized with pre-trained weights from ImageNet1K \cite{deng2009imagenet}. 
We adopt SGD for optimization and the momentum and weight decay are set to 0.9 and 0.0005.
There are 16 labeled images and 16 unlabeled images in a mini-batch.
The learning rate decays as $lr = lr_{base} \times (1-\frac{iter}{max\_iter})^{0.9}$ at every iteration.
We utilize random resize, random crop, and random horizontal as weak augmentations.
Color jitter, random grayscale, random gaussian blur, and CutMix \cite{yun2019cutmix} are employed as strong augmentations. 
Empirically, $p$ and $\gamma$ are set to 32 and 0.7, $\sigma_1$ and $\sigma_2$ are set to 0.7 and 1.5, $\lambda_1$ and $\lambda_2$ are set to 1, $\tau$ is set to 0.90.
For \textit{Pascal VOC 2012}/\textit{Cityscapes}, we crop images into $513 \times 513$/$769\times769$, and train 80/240 epochs with a base learning rate of 0.001/0.01.
The OHEM loss \cite{shrivastava2016training} is applied for \textit{Cityscapes}.

\subsection{Main Results}
\label{sec:results}

\noindent\textbf{Results on \textit{Pascal VOC 2012} classic partitions.}
Table \ref{tab:pascal_classic} presents the results for classic partitions on \textit{Pascal VOC 2012}, where all methods are trained with the CPS partition protocols equipped with the ResNet-101 backbone. 
Compare to supervised-only baseline, \texttt{MaskMatch} shows improvements of +35.36\%, +19.63\%, +13.96\%, +10.09\% and +7.34\% under 1/16, 1/8, 1/4, 1/2 and full partitions, respectively. 
Moreover, \texttt{MaskMatch} outperforms the state-of-the-art approach PCR \cite{xu2022semi} by +5.56\%, +1.40\%, +2.22\%, +2.20\% and +1.14\% under 1/16, 1/8, 1/4, 1/2 and full partitions, respectively. 
Notably, due to its effectiveness in capturing fine-grained details, our approach outperforms the previous state-of-the-art in all partitions, especially when only a few labeled images are available.

\begin{table}[h] 
\caption{
Results on \textit{Pascal VOC 2012} validation set under various \textit{classic} partition protocols, obtained with the ResNet-101 backbone following previous baselines.
The labeled images are selected from the original high-quality set which comprises a total of 1464 labeled images.
The fractions indicate the proportion of labeled images in the original high-quality set, followed by the exact quantity of labeled images.
``Full" denotes that the entire original high-quality set is utilized as the labeled set.
``SupOnly" indicates the supervised-only baseline without any unlabeled images for training.}
\label{tab:pascal_classic}
\centering
\resizebox{\linewidth}{!}{
\begin{tabular}{l|ccccc}
\hline\hline
{Method} & { 1/16 (92) }  & { 1/8 (183) }  & { 1/4 (366) }  & { 1/2 (732) } & {Full (1464)} \\ \hline
{SupOnly} & {40.26} & {56.48} & {65.42} & {70.60} & {74.45}  \\ \hline
{PseudoSeg \cite{zou2020pseudoseg} {\tiny\textcolor{gray} {[ICLR'21]}}} & {57.60} & {65.50} & {69.14} & {72.41} & {/} \\ 
{CPS \cite{chen2021semi} {\tiny\textcolor{gray} {[CVPR'21]}}} & {64.07} & {67.42} & {71.71} & {75.88} & {/}\\ 
{PSMT \cite{liu2022perturbed} {\tiny\textcolor{gray} {[CVPR'22]}}}  & {65.80} & {69.58} & {76.57} & {78.42} & {/}  \\ 
{ST++ \cite{yang2022st++} {\tiny\textcolor{gray} {[CVPR'22]}}}   & {65.20} & {71.00} & {74.60} & {77.30} & {79.10}  \\ 
{U$^2$PL \cite{wang2022semi} {\tiny\textcolor{gray} {[CVPR'22]}}}  & {67.98} & {69.15} & {73.66} & {76.16} & {79.49}  \\ 
{GTA-Seg \cite{jin2022semi} {\tiny\textcolor{gray} {[NeurIPS'22]}}} & {70.02} & {73.16} & {75.57} & {78.37} & {80.47} \\
{PCR \cite{xu2022semi} {\tiny\textcolor{gray} {[NeurIPS'22]}}} & {70.06} & {74.71} & {77.16} & {78.49} & {80.65}  \\ \hline 
{MaskMatch (ours)} &  \textbf{75.62} & \textbf{76.11} & \textbf{79.38} & \textbf{80.69} & \textbf{81.79} \\ \hline \hline
\end{tabular}
}
\end{table}

\noindent\textbf{Results on \textit{Pascal VOC 2012} blender partitions.}
Table \ref{tab:pascal_blender} summarizes the results for blender partitions on \textit{Pascal VOC 2012}.
CPS \cite{chen2021semi}, PSMT \cite{liu2022perturbed} and ST++ \cite{yang2022st++} are trained under the CPS partition protocols. 
Following them, experiments with the ResNet-50 backbone are additionally conducted. AEL \cite{hu2021semi}, U$^2$PL \cite{wang2022semi} , PCR \cite{xu2022semi} and UniMatch \cite{unimatch} are trained under the U$^2$PL partition protocols. 
Following them, the experiments are only based on the ResNet-101 backbone for consistent comparison. 
For CPS blender partition protocols, our method surpasses the existing state-of-the-art method PSMT \cite{liu2022perturbed} by +2.54\%, +1.12\%, and +0.97\% under 1/16, 1/8, and 1/4 partitions with ResNet-50, respectively. 
When enhanced with more powerful backbone ResNet-101, our method outperforms previous SOTA methods by +1.16\%, +0.36\%, +0.72\% under 1/16, 1/8, 1/4 partitions, respectively.
As for the U$^2$PL blender partition protocols, our method still achieves superior performance compare to previous SOTA with gains of +0.01\%, +0.66\% and 0.20\% under the 1/16, 1/8 and 1/4 protocols, respectively.

\begin{table*}[ht]
\caption{
Results on \textit{Pascal VOC 2012} validation set under various \textit{blender} partition protocols. 
The labeled images are selected from augmented \textit{Pascal VOC 2012} which consists of 10482 labeled images. 
$\dagger$ means the results are obtained under the CPS partition.
$\ddagger$ indicates the results obtained under the U$^2$PL partition. Note that under the U$^2$PL blender partition, more labeled images are used, more noisy annotations for supervision.
}
\label{tab:pascal_blender}
\centering
\resizebox{0.8\linewidth}{!}{
\begin{tabular}{l|ccc|ccc}
\hline\hline 
\multirow{2}*{Method} & \multicolumn{3}{c|}{{ ResNet-50 }}  & \multicolumn{3}{c}{{ ResNet-101}}  \\ \cline{2-7} 
{} & { 1/16 (662) }  & { 1/8 (1323) }  & { 1/4 (2646) }  & { 1/16 (662) }  & { 1/8 (1323) }  & { 1/4 (2646) }  \\ \hline \hline
{SupOnly} & {59.82} & {66.53} & {71.64} & {67.03} & {72.15} & {74.52}\\ \hline
{MT \cite{tarvainen2017mean}$\dagger$ {\tiny\textcolor{gray} {[ICLR'17]}}} & {66.77} & {70.78} & {73.22} & {70.59} & {73.20} & {76.62}\\ 
{CCT \cite{ouali2020semi}$\dagger$ {\tiny\textcolor{gray} {[CVPR'20]}}} & {65.22} & {70.87} & {73.43} & {67.94} & {73.00} & {76.17}\\ 
{CutMix-Seg \cite{french2019semi}$\dagger$ {\tiny\textcolor{gray} {[BMVC'20]}}} & {68.90} & {70.70} & {72.46} & {72.56} & {72.69} & {74.25} \\
{GCT \cite{ke2020guided}$\dagger$ {\tiny\textcolor{gray} {[ECCV'20]}}} & {64.05} & {70.47} & {73.45} & {69.77} & {73.30} & {75.25} \\ 
{CPS \cite{chen2021semi}$\dagger$ {\tiny\textcolor{gray} {[CVPR'21]}}} & {71.98} & {73.67} & {74.90} & {74.48} & {76.44} & {77.68} \\ 
{PSMT \cite{liu2022perturbed}$\dagger$ {\tiny\textcolor{gray} {[CVPR'22]}}} & {72.83} & {75.70} & {76.43} & {75.50} & {78.20} & {78.72}\\ \hline
{MaskMatch (ours)$\dagger$} &  \textbf{75.37} & \textbf{76.82} & \textbf{77.40} &  \textbf{76.66} & \textbf{78.56} & \textbf{79.44} \\ \hline \hline
{AEL \cite{hu2021semi}$\ddagger$ {\tiny\textcolor{gray} {[NeurIPS'21]}}} & {/} & {/} & {/} & {77.20} & {77.57} & {78.06}\\ 
{U$^2$PL \cite{wang2022semi}$\ddagger$ {\tiny\textcolor{gray} {[CVPR'22]}}} & {/} & {/} & {/} & {77.21} & {79.01} & {79.30}\\ 
{GTA-Seg \cite{jin2022semi} {\tiny\textcolor{gray} {[NeurIPS'22]}}}& {/} & {/} & {/} & {77.82} & {80.47} & {80.57}\\
{PCR \cite{xu2022semi}$\ddagger$ {\tiny\textcolor{gray} {[NeurIPS'22]}}}& {/} & {/} & {/} & {78.60} & {80.71} & {80.78}\\ 
{UniMatch\cite{unimatch}  {\tiny\textcolor{gray} {[CVPR'23]}}} & {/} & {/} & {/} &  {80.94} & {81.92} & {80.43}\\ \hline
{MaskMatch (ours)$\ddagger$} &  {/} & {/} & {/} & \textbf{80.95} & \textbf{82.58} & \textbf{80.98}\\ \hline \hline 
\end{tabular}
}
\end{table*}

\noindent\textbf{Results on \textit{Cityscapes}.}
The ResNet-101 backbone based results on \textit{Cityscapes} are reported in Table \ref{tab:cityscapes}.
Compared to the supervised-only baseline, our method shows remarkable improvements of +10.66\%, +5.34\%, +3.31\%, +3.61\% under the 1/16, 1/8, 1/4, 1/2 partition protocols.
Our method outperform previous SOTA UniMatch \cite{unimatch} by +0.21\%, +1.11\%, +1.21\% under the 1/8, 1/4, 1/2 partition protocols, and achieve a comparable performance under the 1/16 partition protocol.

\begin{table}[h]
\caption{
Results on \textit{Cityscapes} validation set under various partition protocols, obtained with the ResNet-101 backbone following previous baselines.
$\dagger$ represent the results are reported by PSMT.
$*$: we reproduce the results with the training resolution of 769.
}
\label{tab:cityscapes}
\centering
\resizebox{\linewidth}{!}{
\begin{tabular}{l|cccc}
\hline\hline 
{Method} & { 1/16 (186) }  & { 1/8 (372) }  & { 1/4 (744) }  & { 1/2 (1488) }  \\ \hline \hline 
{SupOnly} & {65.02} & {72.48 } & {75.40 } & {76.68}\\ \hline
{MT \cite{tarvainen2017mean}$\dagger$ {\tiny\textcolor{gray} {[ICLR'17]}}} & {68.08} & {73.71} & {76.53} & {78.59} \\ 
{CCT \cite{ouali2020semi}$\dagger$ {\tiny\textcolor{gray} {[CVPR'20]}}} & {69.64} & {74.48} & {76.35} & {78.29} \\ 
{GCT \cite{ke2020guided}$\dagger$ {\tiny\textcolor{gray} {[ECCV'20]}}}  & {66.90} & {72.96} & {76.45} & {78.58} \\ 
{CPS \cite{chen2021semi}$\dagger$ {\tiny\textcolor{gray} {[CVPR'21]}}} & {74.48} & {76.44} & {77.68} & {78.64} \\
{PSMT \cite{liu2022perturbed} {\tiny\textcolor{gray} {[CVPR'22]}}} & {/} & {76.89} & {77.60} & {79.09} \\  
{AEL \cite{hu2021semi} {\tiny\textcolor{gray} {[NeurIPS'21]}}}  & {74.45} & {75.55} & {77.48} & {79.01} \\ 
{U$^2$PL \cite{wang2022semi} {\tiny\textcolor{gray} {[CVPR'22]}}} & {70.30} & {74.37} & {76.47} & {79.05} \\ 
{PCR \cite{xu2022semi} {\tiny\textcolor{gray} {[NeurIPS'22]}}}& {73.41} & {76.31} & {78.40} & {79.11}  \\  {UniMatch\cite{unimatch}$*$  {\tiny\textcolor{gray} {[CVPR'23]}}} & \textbf{75.76} & {77.61} & {77.60} & {79.08} \\ \hline 
{MaskMatch (ours)} &  {75.68} & \textbf{77.82} & \textbf{78.71} & \textbf{80.29} \\ \hline \hline 
\end{tabular}
}
\end{table}

\noindent\textbf{Performance along Boundaries.}
As a part of fine-grained local semantics, we further evaluate the performance of boundary predictions.
Following \cite{Ma2021Boundary}, we adapt the boundary F-score with threshold of 0.0003 to measure the performance of boundary regions.
Results on Cityscapes validation set under various partition protocols are presented in Table \ref{tab:results_boundary}.
Our \texttt{MaskMatch} surpasses previous SOTA method UniMatch \cite{unimatch} by +0.53\%, +0.63\%, +1.37\% and +0.38\% on the boundary F-score under the 1/16, 1/8, 1/4 and 1/2 partition protocols.
Despite the slightly lower mIOU exhibited by our method compared to UniMatch within the 1/16 partition protocol, our method still outperforms UniMatch in terms of the boundary F-score (Table \ref{tab:results_boundary}).
Results demonstrate the superiority of our method in learning fine-grained locality.

\begin{table}[h]
\caption{
Comparison of the Boundary F-score on \textit{Cityscapes} validation set under various partition protocols. 
}
\label{tab:results_boundary}
\centering
\resizebox{0.97\linewidth}{!}{
\begin{tabular}{l|cccc}
\hline \hline
{Method} & { 1/16 (186) }  & { 1/8 (372) }  & { 1/4 (744) }  & { 1/2 (1488) }  \\ \hline
{UniMatch\cite{unimatch}  {\tiny\textcolor{gray} {[CVPR'23]}}} & {56.28} & {57.35} & {57.71} & {58.76} \\ 
{MaskMatch (ours)} &  \textbf{56.81} & \textbf{57.98} & \textbf{59.08} & \textbf{59.14} \\ \hline \hline 
\end{tabular}
}
\end{table}

\noindent \textbf{Comparison of FLOPs.}
The comparison of FLOPs is presented in Table \ref{tab:flops}. Upon incorporating LCR, our approach achieves superior performance (81.31\%) with lower GFLOPs (2058.41). 
When LCR and MS are combined, our approach entails slightly higher computational costs compared to other methods. 
Nonetheless, the added computational burden is deemed acceptable due to the remarkable effectiveness of our approach.

\begin{table}[h]
\caption{
Comparison of computational cost with other methods.
mIOU on \textit{Pascal VOC 2012} under U$^2$PL blender 1/8 partition is reported.
}
\label{tab:flops}
\centering
\resizebox{0.97\linewidth}{!}{
\begin{tabular}[\linewidth]{l|ccccccc}
\hline \hline 
{Method} & {CPS} & {PSMT} & {U$^2$PL} & {GTA-Seg} & {UniMatch} & {MaskMatch \tiny{w/o MS}} & {MaskMatch}\\ \hline
{GFLOPs} & {4119.86}  & {2177.71} & {2987.00}  & {4091.21} & {2891.15} &  \textbf{2058.41} & {4844.49}\\
{mIOU(\%)} & {76.44}  & {78.20} & {79.01}  & {80.47} & {81.92} & {81.35}  & \textbf{82.58}\\\hline\hline 
\end{tabular}
}
\end{table}

\noindent \textbf{Plug-and-play LCR.}
The proposed LCR does not introduce extra parameters or modify the backbone, ensuring plug-and-play flexibility. We integrate LCR into U$^2$PL. 
As illustrated in Table \ref{tab:plug_and_play}, we observe an improvement of +1.34\%, proving the plug-and-play capacity of LCR. 

\begin{table}[h]
\caption{
    Results of plugging LCR into U$^2$PL on \textit{Pascal VOC 2012} under classic full partition.
    Due to GPU memory constraints, we reproduce the results using a batch size of 8.
}
\label{tab:plug_and_play}
\centering
\resizebox{1.0\linewidth}{!}{
\begin{tabular}[\linewidth]{lcccc}
\hline \hline
{Method} & {Paper Reported} & {Our Reproduced} & {Reproduced w/ LCR}  & {$\Delta$} \\ \hline
{U$^2$PL} & {79.49} & {78.67} & {80.01} & {+1.34}\\
\hline \hline
\end{tabular}
}
\end{table}

\begin{figure}[h]
\begin{center}
\includegraphics[width=0.90\linewidth]{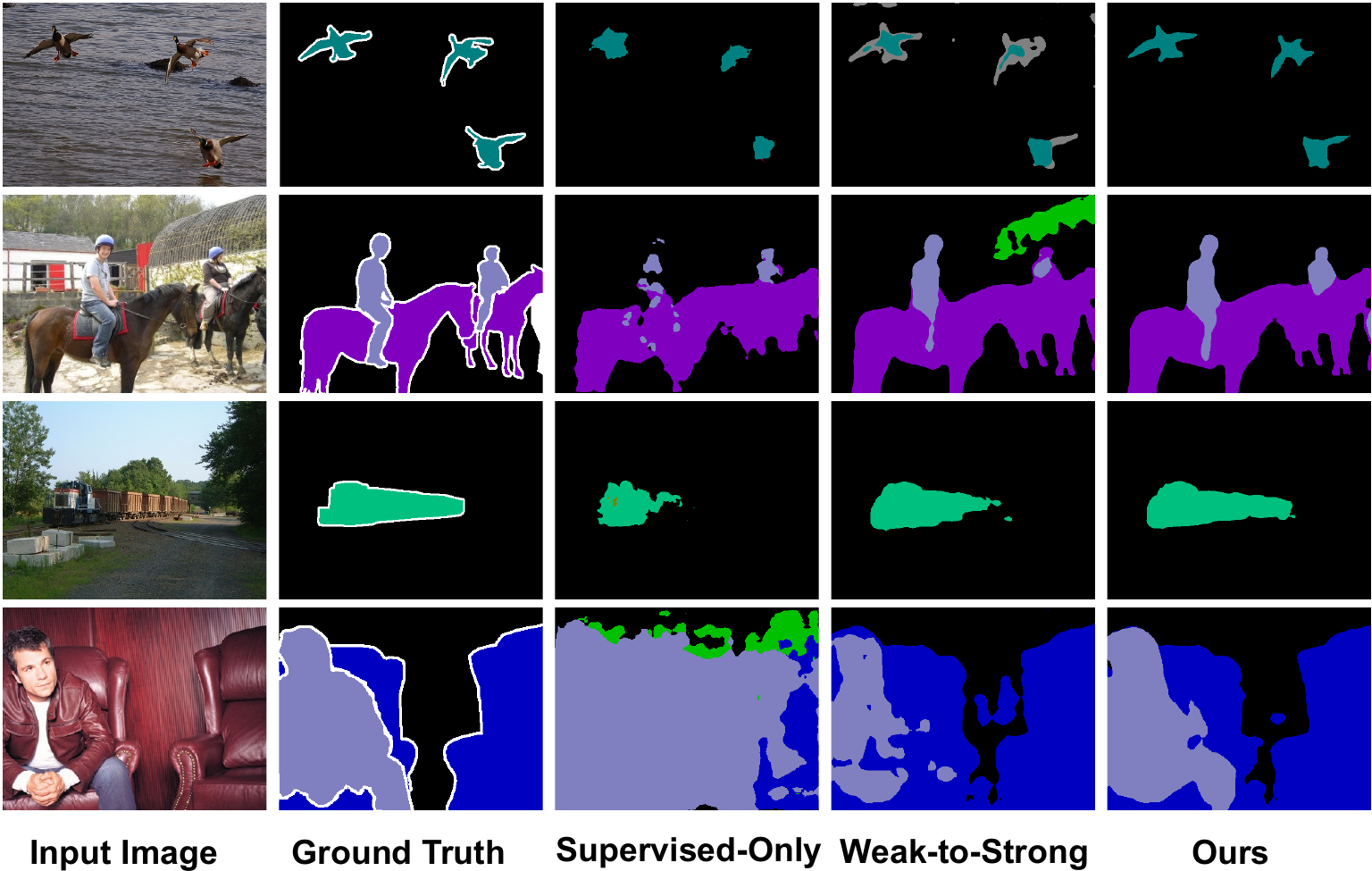}
\end{center}
   \caption{
Qualitative results on \textit{Pascal VOC 2012} validation set under 1/16 classic partition protocol.
}
\label{fig:visulization}
\vspace{-1.0em}
\end{figure}

\noindent\textbf{Visualization.}
Figure \ref{fig:visulization} presents the qualitative visual results on \textit{Pascal VOC 2012} validation set with only 92 labeled samples.
As observed, the supervised-only manner, which learns from limited labeled data only, appears to be vulnerable with poor generalizability, especially for the cases at the 3rd and 4th rows.
In contrast, clearly stronger performance can be achieved when the abundant unlabeled data is utilized (the baseline that incorporating weak-to-strong consistency regularization and mean teacher paradigm and our method). 
Compared to the baseline, our proposed approach shows a stronger ability in recognizing the correct categories (the 1st and 2nd row) and superior performance in determining fine-grained local contours (the 3rd and 4th row). 

\subsection{Ablation study}

\noindent\textbf{Effectiveness of each component.}
As presented in Table \ref{tab:ablation}, we construct \texttt{MaskMatch} step-by-step to evaluate the effectiveness of each module on \textit{Pascal VOC 2012} under the 1/8 partition U$^2$PL setting and \textit{Cityscapes} under the 1/8 partition setting. 
All experiments are conducted with ResNet101 backbone.
The mean teacher baseline model incorporated weak-to-strong consistency and pseudo label selection achieves a segmentation mIoU of 80.24\% and 76.45\%.
MS achieves an improvement of 0.66\% and 0.22\%.
LCR yields 1.15\% and 1.16\% improvement, showing the effectiveness of our proposed auxiliary masked modeling proxy task.
\texttt{MaskMatch}, incorporating two strategies for fine-grained locality learning, achieves a more appealing improvement of 2.34\% and 1.37\%.

We also provide the Boundary F-score in Table \ref{tab:ablation} to show the impact of each component on boundary regions.
The integration of LCR yields a notable improvement of +0.52\%, demonstrating its effectiveness in improving local regions.
Incorporating MS only bring insignificant performance gains +0.05\%. However, when MS is combined with LCR, the performance is boosted by +1.37\%.
The improvement of locality perception mainly stems from LCR and the collaboration (LCR+MS) instead of MS alone.

\begin{table}[h]
\caption{
Ablation study: mean teacher paradigms (MT), weak-to-strong consistency (WS), pseudo label selection (PS), multi-scale strategy (MS), and local consistency regularization (LCR).
}
\label{tab:ablation}
\centering
\resizebox{0.95\linewidth}{!}{
\begin{tabular}{ccccc|c|cc}
\hline\hline 
\multirow{2}*{MT} & \multirow{2}*{WS} & \multirow{2}*{PS} & \multirow{2}*{MS} & \multirow{2}*{LCR} & { Pascal(1323) }  & \multicolumn{2}{c}{{ Cityscapes(372)}}  \\ \cline{6-8} 
{} & {}  & {}  & {}  & {}  & {mIOU(\%)}  & {mIOU(\%)}  & {F-score(Boundary) (\%)}  \\ \hline 
{$\checkmark$} & {$\checkmark$} & {$\checkmark$} & {} & {} & {80.24} & {76.45} & {56.61}\\ 
{$\checkmark$} & {$\checkmark$} & {$\checkmark$} & {$\checkmark$} & {} & {80.90}  & {76.67} & {56.66}\\ 
{$\checkmark$} & {$\checkmark$} & {$\checkmark$} & {} & {$\checkmark$} & {81.35}  & {77.61} & {57.13}\\  
{$\checkmark$} & {$\checkmark$} & {$\checkmark$} & {$\checkmark$} & {$\checkmark$} & {82.58} & {77.82} & {57.98}\\  \hline \hline 
\end{tabular}
}
\end{table}

\noindent\textbf{Analysis on hyper-parameters.}
See supplementary material for more analysis on patch size $p$, mask ratio $\gamma$, pseudo-label threshold $\tau$ and the trade-off weights ($\lambda_1$, $\lambda_2$).

\begin{figure}[h]
\begin{center}
\includegraphics[width=0.8\linewidth]{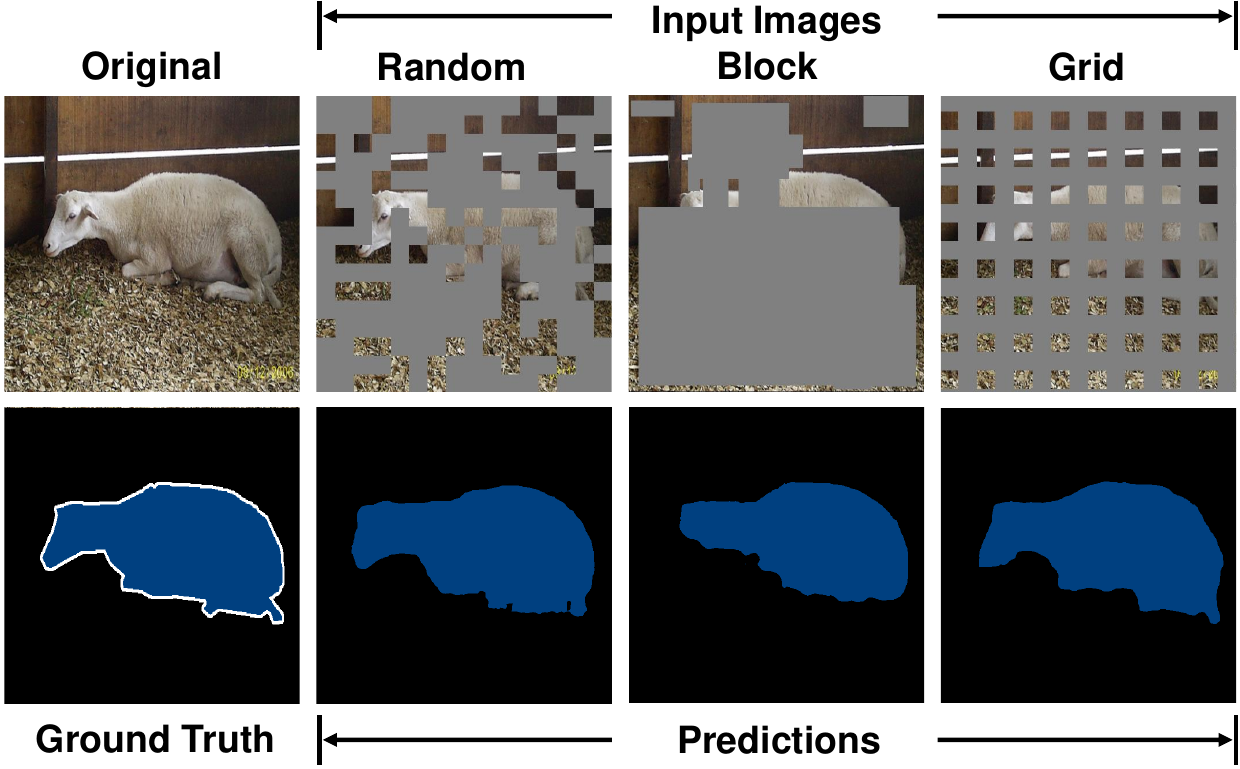}
\end{center}
   \caption{
   Comparison of different mask sampling strategies. 
   The original image and ground truth are placed on the left. 
   The masked input images (row 1) and corresponding predictions (row 2) of random (\texttt{MaskMatch}), block-wise \cite{bao2022beit}, and grid sampling strategy are displayed from the left 2nd column to the right. 
   The masked patches are colored in gray. Our model learns to predict segmentation masks even with limited unmasked local patches.
}
\label{fig:mask_sampling_strategies}
\vspace{-1.0em}
\end{figure}

\noindent\textbf{Comparison on Different Masked Image Modeling Strategies.}
We further delve into the mask-then-predict paradigm and investigate two prediction tasks: our segmentation label prediction task and vanilla image reconstruction task similar to MIM \cite{he2022masked}.
Note that since there are masked and unmasked regions, we explore to integrate different tasks into different regions. Particularly,
an auxiliary reconstruction head is added to perform image reconstruction. Table \ref{tab:dissusion_task} shows that incorporating the label prediction task on the entire image is the most effective way to enhance semi-supervised semantic segmentation. 
Interestingly, there is a decrease in performance comparing to our current \texttt{MaskMatch} if we only require the model to perform segmentation on the unmasked patches. 
We attribute this to the fact that directly predicting label on the masked patches (see Figure \ref{fig:mask_sampling_strategies}) forces the model reconstructing semantics on masked patches by leveraging information from near unmasked patches, which further encourages the model to learn fine-grained and discriminative local semantics.
However, the mask-then-reconstruct approach shows poor performance, even compared with the baseline. 
We believe this is due to the incompatibility of high-level semantic segmentation and low-level image reconstruction, where the model struggles to learn a suitable representation for both dense prediction tasks. 
Similar findings can be found in \cite{xie2022simmim, he2022masked}, where the linear evaluation performance of the MIM pre-trained backbone is inferior to those of backbones pre-trained using semantic based contrastive learning approaches \cite{chen2021mocov3, caron2021emerging}.
Overall, we contend that the task-specific design in \texttt{MaskMatch} is well-suited for semi-supervised segmentation to meet the spirit of masked modeling.

\begin{table}[h]
\caption{
Results of different potential masked image modeling strategy on 1/8 blender partition protocol and 1/16 classic partition protocol. 
After separating images into masked regions and unmasked regions, 
tasks include image reconstruction (IR) and label prediction (LP) are performed on these regions independently. 
'/' indicates no tasks are performed on the targeted regions.
}
\label{tab:dissusion_task}
\centering
\resizebox{0.9\linewidth}{!}{
\begin{tabular}[\linewidth]{c|c|cc}
\hline\hline 
\multicolumn{2}{c|}{{ Task }} & \multirow{2}*{blender (1323)} & \multirow{2}*{classic (92)}\\ \cline{1-2}
{Unmasked Regions} & {Masked Regions} & {} & {} \\ \hline
{/} & {/} & {80.90} & {72.62} \\
{LP} & {/} & {81.96} & {74.48}  \\
{LP} & {LP} & {\textbf{82.58}} & {\textbf{75.63}} \\
{IR} & {IR} & {76.98} & {63.74} \\
\hline  \hline 
\end{tabular}
}
\end{table}

\noindent\textbf{Comparison on Different Mask Sampling Strategies.}
Table \ref{tab:dissusion_mask_sampling_strategies} summarizes how different mask sampling strategies (illustrated in Figure \ref{fig:mask_sampling_strategies}) affect the segmentation results. 
As observed, the random sampling strategy outperforms the block-wise and the grid-wise sampling strategy. As shown in Figure \ref{fig:mask_sampling_strategies}, the block-wise sampling strategy \cite{bao2022beit} tends to eliminate large blocks, which fails to provide enough individual patches for fine-grained locality learning. 
The grid-wise sampling strategy, which preserves one out of every four patches, provides more information for masked prediction and making such task easier.
However, it have a lower ability to capture fine-grained locality. 
Overall, the random sampling strategy in \texttt{MaskMatch} is more effective.

\begin{table}[h]
\caption{Results of various mask sampling strategies with masked image approach on 1/8 blender partition protocol and 1/16 classic partition protocol, respectively.}
\label{tab:dissusion_mask_sampling_strategies}
\centering
\resizebox{0.68\linewidth}{!}{\begin{tabular}[\linewidth]{c|ccc}
\hline \hline 
{Strategies} & {Random} & {Block} & {Grid
}\\ \hline
{blender (1323)} & \textbf{82.58} & {81.67} & {81.82} \\ 
{classic (92)} & \textbf{75.62} & {74.02} & {74.66} \\ \hline\hline
\end{tabular}}
\end{table}

\section{Conclusion}
In this paper, we proposed \texttt{MaskMatch} to address the inherent neglect of local semantic patterns in previous consistency regularization based semi-supervised semantic segmentation methods.
Our framework introduces a task-specific auxiliary masked modeling proxy task to facilitate fine-grained locality learning assisted by a multi-scale ensembling strategy for high-quality pseudo-label generation.
Extensive experiments on two public benchmark datasets show the superior performance of our method.

{\small
\bibliographystyle{ieee_fullname}
\bibliography{egbib}
}

\end{document}